\documentclass{article}
\usepackage{arxiv}
\usepackage{booktabs}
\usepackage{amsfonts}
\usepackage{graphicx}
\usepackage{comment}
\usepackage{subcaption}
\usepackage{multirow}
\usepackage{multicol}
\usepackage{xcolor}
\usepackage{url}
\usepackage{setspace}
\usepackage{amsmath}
\usepackage{lineno}

\title{A data-centric weak supervised learning for highway traffic incident detection}

\author{
  Yixuan Sun \\
  School of Mechanical Engineering, \\
  Purdue University,  West Lafayette, IN \\
  \texttt{yixuan-sun@purdue.edu} \\
   \And
  Tanwi Mallick \\ 
  Mathematics and Computer Science Division\\
  Argonne National Laboratory, Lemont, IL \\
  \texttt{tmallick@anl.gov} \\
   \And
  Prasanna Balaprakash \\
  Mathematics and Computer Science Division \& 
  Argonne Leadership Computing Facility \\
  Argonne National Laboratory, Lemont, IL \\
  \texttt{pbalapra@anl.gov} \\
  
  \And
  Jane Macfarlane\\
  Sustainable Energy Systems Group \\
  Lawrence Berkeley National Laboratory, Berkeley, CA \\
  \texttt{jfmacfarlane@lbl.gov} \\
}

\begin{document}

\maketitle

\begin{abstract}

Using the data from loop detector sensors for near-real-time detection of traffic incidents in highways  is crucial to averting major traffic congestion.      
While recent supervised machine learning methods offer solutions to incident detection by leveraging  human-labeled incident data, the false alarm rate is often too high to be used in practice. Specifically, the inconsistency in the human labeling of the incidents significantly affects the performance of supervised learning models. To that end, we focus on a data-centric approach to improve the accuracy and reduce the false alarm rate of the traffic incident detection on highways. We develop a weak supervised learning  workflow to generate high-quality training labels for the incident data without the ground truth labels, and we use those generated labels in the supervised learning setup for final detection. This approach comprises three stages. First, we introduce a data preprocessing and curation pipeline that processes traffic sensor data to generate high-quality training data through leveraging labeling functions, which can be domain knowledge related or simple heuristic rules. Second, we evaluate the training data generated by weak supervision using three supervised learning models---random forest, k-nearest neighbors,  and a support vector machine  ensemble---and long short-term memory classifiers. The results show that  the accuracy of all of the models  improves significantly after using the training data generated by weak supervision. 
Third, we develop an online real-time incident detection approach that leverages the model ensemble and the uncertainty quantification while detecting incidents. Overall, we show that our proposed weak supervised learning workflow achieves a high incident detection rate (0.90) and low false alarm rate (0.08).

\end{abstract}

\section{Introduction}

Traffic congestion negatively impacts quality of life and economic productivity, causing longer travel times and loss of productivity \cite{falcocchio2015costs}. According to the U.S. Department of Transportation \cite{systematics2005traffic} 
25\% of traffic congestion stems from the direct consequence of traffic incidents. Moreover, traffic incidents can cause secondary incidents that lead to additional property loss, injury, and even death. Therefore, a fast and reliable traffic incident detection system is needed to detect incidents quickly and accurately so that traffic management controllers can make adjustments in real time to redirect traffic. Achieving accurate automated incident detection (AID) has been a research topic for decades. With the help of increasingly abundant data describing traffic conditions, data-driven methods for incident detection have been developed. These methods seek to provide fast and accurate traffic incident detection with a high detection rate (DR) and low false alarm rate (FAR). Specifically, methods based on time series analysis and machine learning (ML) have  become the most popular for their ability to explore the temporal relationships among traffic variables and the potential of utilizing hidden patterns within the traffic data.

Real-world traffic data are  noisy and  challenging to curate for pattern extraction. Traffic patterns can vary with different road conditions, weather, and time of the day, making detection of incidents difficult.  The most commonly used data for data-driven incident detection are inductive loop detector measurements. Not only are the measurement values  dependent on the detector condition, but also their use must account for localized road geometry (e.g., number of lanes, entrances, exits, intersections). Moreover, the distance between adjacent loop detectors is not uniform, a situation that, depending on the distance between the detectors, can result in pattern shifts between detectors. For incident detection tasks in particular, the success of data-driven models  greatly depends on the reported incident events, which do not always match well with the observed traffic condition.
Therefore, careful treatment of the real-world traffic data is required for the data to be useful for AID.  

Previous studies on AID using ML methods  focused mainly on using and improving different models to achieve better detection. However, the impact of the quality of the data  used to train the models was overlooked---even though the correctness of the training data fundamentally determines the performance of a data-driven model, regardless of the model complexity. 
In recent ML approaches, data quality has gained increasing attention  as  a major driving force for a successful machine learning model. Researchers are recognizing that high-quality training data is required in order to perform prediction with high accuracy \cite{gudivada2017data}. 

In this paper we introduce a data-centric workflow that processes raw incident data collected from the Performance Measurement System (PeMS) database \cite{chen2002freeway}, and we generate high-quality training data using a weak supervision technique. The data generated by this technique are further utilized by a number of machine learning models as the high-quality training set. Our evaluation shows that the high-quality training data improve the model accuracy significantly and greatly reduce the FAR introduced by seemingly 
incident-related traffic patterns during normal traffic conditions. Further, we  perform uncertainty quantification for AID using deep ensemble. 
As the first work on incident detection using the data-centric approach, the contributions of this study are as follows.
(1) Data-centric modeling is emphasized.  High-quality data are generated via weak supervision. The training set created by weak supervision requires no feature engineering. 
(2) The importance of data preprocessing and curation is discussed and presented by comparing the performance of models with the same architecture trained on different levels of treated data. 
(3) The importance of the quality of the training data on the model performance is shown by evaluating the training data generated via weak supervision using three machine learning models---random forest, k-nearest neighbors (KNN), and support vector machine (SVM) ensemble---and long short-term memory (LSTM) classifiers. To the best of our knowledge, this is the first application of LSTM to detector-level traffic incident detection with only loop detector measurements. 
(4) Our uncertainty quantification for AID is helpful for understanding whether the model prediction is trustworthy or not.

\begin{figure}[t]
    \centering
    \includegraphics[width=0.7\linewidth]{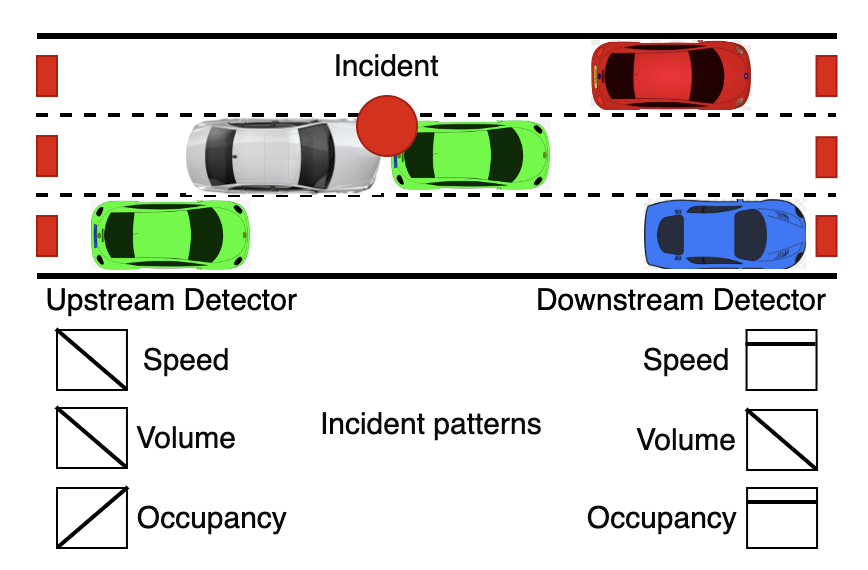}
    \caption{Traffic incident pattern schematic (\cite{motamed2016developing}). During a traffic incident, the corresponding upstream and downstream loop detectors will measure different traffic patterns. The expected response is a decrease in upstream speed and volume and in downstream volume, with an increase in upstream occupancy and changed downstream speed and occupancy. 
    }
    \label{schematic}
\end{figure}

\section{Literature review}
\label{related}
Automatic incident detection on traffic networks has been a research topic for decades. Starting in the 1970s, algorithms were developed to detect potential incidents on the traffic  networks. These AID algorithms were  based mostly on the information provided by the inductive loop detectors embedded in the freeway pavement, from which traffic occupancy and volume profiles could be recorded. Accordingly, with the usage of occupancy, the California algorithm \cite{levin1979incident} and its variations were developed. With the installations of dual-loop detectors that can record traffic speed information, the detecting methods were expanded. In general, these algorithms can be categorized into pattern-based algorithms, catastrophe-based methods, and statistical methods \cite{martin2001incident}. With abundant data available from the traffic systems nowadays, data-driven methods, such as ML model-based AID frameworks, have been receiving increased attention.


Traffic data, regardless of the type, can be complicated and noisy because of sophisticated road layouts, human recording of incidents, and faulty sensors. Using ML models to achieve detection from the traffic data has shown great advantages thanks to their ability to learn complex mappings from data. In this section we discuss prior work on using ML models to conduct AID.  While some work has been done on estimating secondary crashes \cite{kitali_likelihood_2018, wang_modeling_2019, park_real-time_2018}, we mainly discuss the studies focusing  on the initial events. 


Several studies focused on detecting whether there is an incident or congestion on a network level, where the detection model takes in the information of the entire traffic network of interest and decides whether an incident is happening within the network without specifying its location. \cite{ma2015large} adopted an RNN combined with a restricted Boltzmann machine to detect network-wide congestion based on GPS data collected from taxis. \cite{zhu2018deep}  developed a CNN-based model to detect incidents in urban traffic networks, where the GIS information was adopted to match the measured traffic parameters and recorded incidents. \cite{liu2019traffic} developed a spatiotemporal graphical model called a spatiotemporal pattern network to predict traffic dynamics in large networks and detect incidents by calculating anomaly scores. \cite{basso_deep_2021} made use of the data from Automatic Vehicle Identification (AVI) gates and devices installed in vehicles, as well as loop detector measurements to detect incident with a CNN model.  In \cite{bao_spatiotemporal_2019},  a spatiotemporal convolutional long short-term memory network (STCL-Net) was used to  predict the city-wide traffic crash risk, where a mixed data sources were used, including crash data, Taxi GPS data, land use, population, and weather information. \cite{huang_highway_2020} transformed the loop detector measurement to heat maps with horizontal axis the time and vertical axis the traffic direction, and applied a CNN model to detect crashes.


As opposed to network-level detection, sensor-level detection emphasizes the usage of local information. \cite{jiang2010automated} developed an AID model where the real-time traffic data were compared with the traffic parameter estimation using a moving average, and then incidents were detected according to the difference. \cite{motamed2016developing} used an SVM  to classify  incident or non-incident instances given traffic speed, flow, and occupancy information. \cite{fouladgar2017scalable} utilized the traffic information on the nodes of interest and their neighbors to construct a spatiotemporal model using CNN and RNN to detect congestion. This work is easily scalable because only the information on a few neighbor nodes is needed. \cite{popescu2017automatic} incorporated vehicle-to-infrastructure communications with the traffic parameters from loop detectors to detect incidents based on distance and time of changing lanes. \cite{singh2018deep} used a stacked autoencoder to generate a high-level representation of traffic surveillance videos and adopted one-class SVM to detect potential traffic accidents. \cite{xiao2019svm} proposed an ensemble of k-nearest neighbors and SVM to classify instances with speed, flow, and occupancy information into incident or non-incident. \cite{yang2020expressway} used CNN and extreme gradient boosting to detect incidents in expressways with data from microwave detectors. \cite{jiang_long_2020} adopted data with different temporal resolution and used  an LSTM-based model to  detect traffic crashes.  \cite{mercader_automatic_2020} proposed to use isolation forest, an unsupervised anomaly detection approach, with bluetooth traffic monitoring data to achieve automatic incident detection.  \cite{parsa_real-time_2019} developed a real-time traffic accident detection model using SVM and probabilistic neural network, and the data imbalance and oversampling were also discussed.  \cite{parsa_toward_2020} adopted different data, e.g., demographic information, land use, and weather features, to detect traffic accident with extreme gradient boosting (XGBoost).  

Network-level detection models rely on the specific traffic network they have been trained on, which cannot be used for a different setting. Thus, we aim to develop a sensor-level detection model so that it has the potential of being deployed at different locations. However, the senor-level detection methods described above  required various types of data,  were evaluated on synthetic data, or did not provide any model prediction uncertainty measure. Also, data quality, which can play an essential role in training effective models, was not mentioned in the previous studies. Our work addresses these issues. Specifically, we adopt a data-centric approach to utilize real-world loop detector speed measurements, curate preprocessed data to create high-quality datasets, develop RNN-based classifiers, and create a deep ensemble model to quantify the model predictive uncertainty.

\section{Dataset}
\label{data}

Loop detector measurements of traffic data and incident data were downloaded from the Caltrans Performance Measurements System (PeMS) database \cite{chen2002freeway}. The data were collected from  detectors spanning the entire California freeway system across all major metropolitan areas. 
From this dataset, a section  of detectors on I-80 East in Caltrans District 4  were arbitrarily selected for analysis. 
The data in the training set spanned 3 months and were sampled from January to March 2019. The validation set came from April and May 2019. The data from the rest of the year 2019 (June--December) were used as the testing set. 
There was no time overlapping between the training, validation, and testing sets. Training and model tuning 
were done on the training and validation set. 
The incident detection was done on the testing set. 

 

The traffic dynamics are captured by inductive loop detectors. Loop detector measurements include speed, volume, and occupancy at a resolution of 30 seconds ($t$ indicates the time step). 
Speed represents the average vehicle speed during a 30-second window for each lane, volume represents the number of vehicles on each lane that passed the detector, and occupancy accounts for the proportion of time during the detecting window that the detector was occupied by the vehicles. 

\begin{figure}
    \centering
    \includegraphics[width=\linewidth]{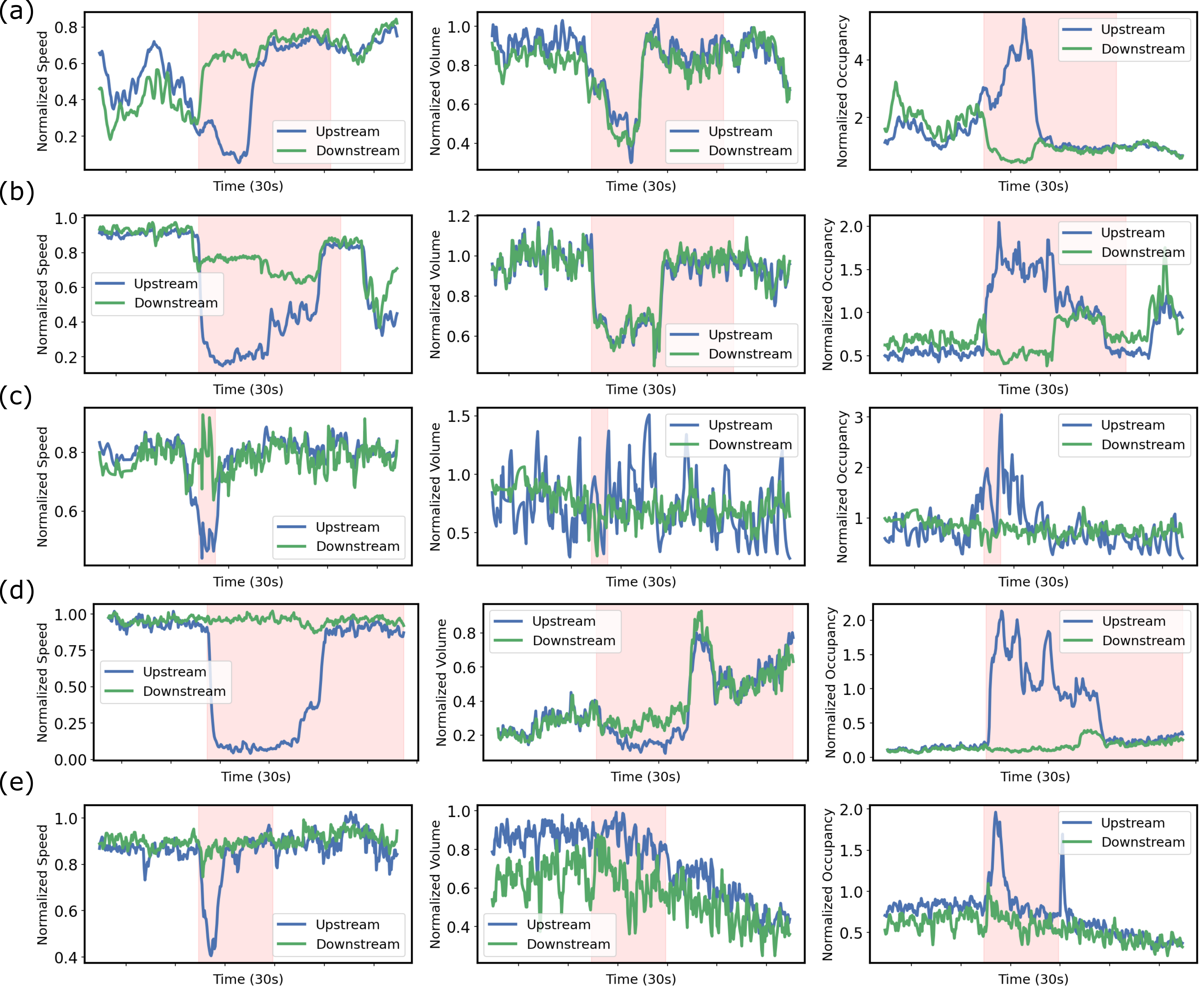}
    \caption{Loop detector measurement changes during an incident, of which the duration is marked in pink background. The expectation is to observe speed and occupancy separation between upstream and downstream detectors and drop in volume from both detectors. More specifically, one expects an upstream speed drop and occupancy increase and a volume increase on both detectors during an incident. (a)(b) match the expectation. The volume profiles in (c)(d)(e) do not follow the expectation, nor do they follow an obvious pattern, while the speed and occupancy profiles still align with the expectation.}
    \label{pattern}
\end{figure}

In order to define an incident, an upstream detector and a downstream detector are established around the incident location. Upstream detector captures the outgoing traffic from the incident location and the downstream detector captures the incoming traffic towards the incident location. 
The expected traffic dynamics include an obvious drop in the incident location's upstream speed while the downstream speed remains the same, Figure~\ref{schematic}.
Although all traffic profiles have their expected pattern changes, in reality they are not always observed, especially for volume change, shown in Figure \ref{pattern}. Thus, in this work we  focus only on utilizing speed and occupancy information to detect incidents. 

\subsection{Data preprocessing}
In order to curate the data for training a data-driven model, the raw data (speed $s^{(t)}$, volume $v^{(t)}$, and occupancy $o^{(t)}$) were further processed and cleaned. The initial step focused on aligning the timestamps of the localized detectors because the detectors operate asynchronously 
Specifically, linear interpolation was used to assign the corresponding values to the desired timestamps. In addition, speed values were reported for each lane ($s_1, s_2, \dots, s_n$); however, the number of lanes $n$ was not constant. Therefore, in  order to reduce the complexity of the problem, the lane information was aggregated as the volume-weighted average of all lanes. 
\begin{linenomath*}
$$ s^{(t)} = \frac{\sum_i^n s_i^{(t)} v_i^{(t)}}{\sum_i^n v_i^{(t)}}$$
\end{linenomath*} 
Furthermore, to deal with missing values, we adopted two different methods. First, for the same loop detector, the missing values were filled with the nearest available value at the previous timestamp. Second, for detectors that had missing values at the first timestamp (e.g., 01-02-2019 00:00), the values were filled with a spline interpolation of spatial locations of consecutive detectors, under the assumption that two adjacent detectors share a similar traffic pattern. 

As with any real-world device, it was not uncommon to see  values that were clearly unrealistic, for example, speed values that were too high for the physical situation. In order to address such situations, an exponential moving average with a step size of 5 (2.5 minutes) was utilized to smooth the 30-second resolution time series data: 
\begin{linenomath*}
$$s^{(t)} \leftarrow \frac{\sum_{j=0}^t w_{j} s^{(t-j)}}{\sum_{j=0}^t w_j}$$
\end{linenomath*}, where $w_j = (1 - \alpha)^j$ is the weight associated to the value at step $j$ ahead of the current time $t$, $\alpha = 0.33$ is the decay factor calculated based on the moving window size 5 (2.5 minutes), and $s^{(t)}$ is the aggregated speed values at time $t$, 
and they are updated by the rule above. The same moving average process was applied to occupancy $o^{(t)}$ and volume $v^{(t)}$.
This approach was helpful for revealing the underlying pattern of the time series data and increasing the robustness of detection models. 

\begin{figure}
    \centering
    \includegraphics[width=0.7\linewidth]{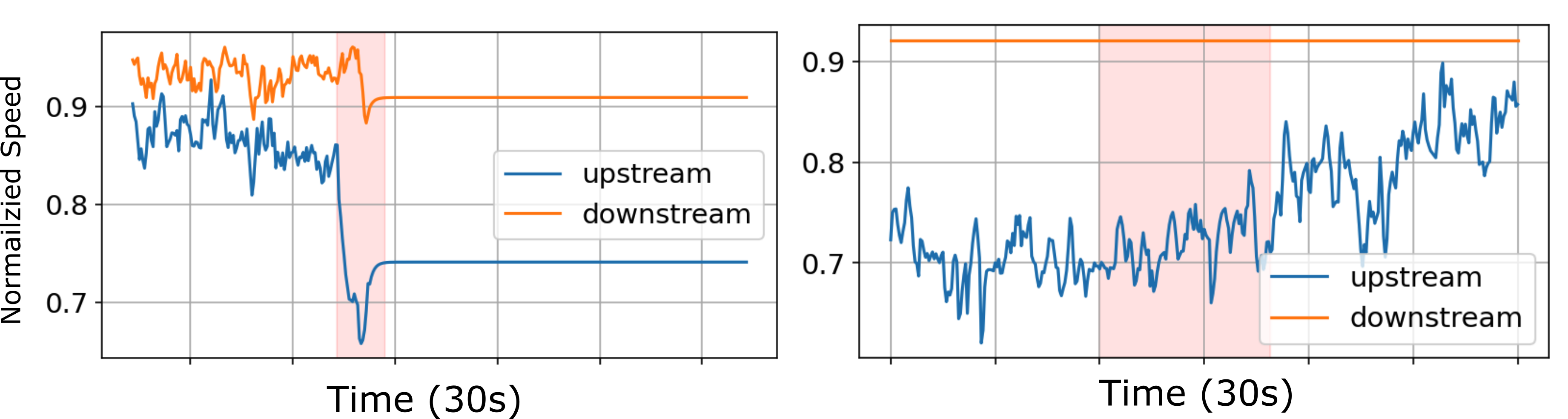}
    \caption{Examples of low-quality data. }
    \label{low_quality}
\end{figure}

Knowing the sensitivity of data-driven models to the quality of the data, we applied an additional filtering step. Because normal data sequences with 30-second resolution usually still exhibit oscillations after smoothing, we assumed that low-quality data would appear as flat lines or extremely smooth curves, and we filtered these data points from the model training and evaluation dataset. This quality check filter was implemented by comparing the mean of the numerical first derivative of each traffic profile to the same statistics from the training set. This method is based on the assumption that the oscillation of normal traffic profiles follows a similar pattern with limited deviation. 
Take speed $s^{(t)}$ as an example. Within a certain time span (e.g., 3 hours), we calculate the numerical first derivative $s'^{(t)}$ and take the mean over the time span as $\bar{s'} = \frac{1}{T}\sum_{t=0}^{T}s'^{(t)}$. This process is repeated for all the examples in the training set. We then take the sample mean and standard deviation of all $\bar{s'},$ 
\begin{linenomath*}
$$\mu_{\bar{s'}} = \frac{\sum_k^m \bar{s'}_{k}}{m},$$ 
$$\sigma_{\bar{s'}} = \frac{\sum_k^m s'_k - \mu_{\bar{s'}}}{m},$$
\end{linenomath*} 
where $m$ is the number of data points in the training set. 
For new traffic profile series, the mean of the numerical first derivative $\bar{s'_{new}}$ will be compared with $\mu_{\bar{s'}} \pm \sigma_{\bar{s'}}$. If $\bar{s'_{new}}$ lands in the range, the new data point is considered to have good qualify; otherwise, it will be considered to have bad quality and will be taken out. 
An example of low-quality data is shown  in Figure \ref{low_quality}, where both data points exhibit data with little variation. The left figure shows a situation in which both upstream and downstream sensors are reporting low-quality data. The right figure indicates that a downstream detector is reporting low-quality data.


\subsection{Labeled incidents}
PeMS includes reports of incidents from traffic management centers or the California Highway Patrol that will provide labels for the associated traffic profiles. Incidents are characterized by incident location, incident starting time and duration, and incident type. Figure \ref{low_quality} includes the reported incident data by showing a shaded bar that indicates the start time and duration of the incidents. Reported incidents in the database are created for a variety of traffic situations, and not all of the reported incidents are captured by loop detector measurements.  For this study, only the incident data that were categorized as accidents and showed evidence of change at the detector level were extracted and matched with loop detector measurements. 

\section{Methodology}
\label{method}

Our approach comprises three key steps:  determine the data curation criterion and process of creating high-quality data for model training, develop a classification model for classifying time slices and the incident detection criteria after classification, and design a real-time detection approach that will fully identify and classify sensor-detectable cases.

In the supervised learning setting, with the input-output pair $\mathbf{x}$ and $Y$, the model input was prepossessed with traffic speed and occupancy time series data 
\begin{linenomath*}
$$\mathbf{x} = (\{s^{(t-h)}, \dots, s^{(t)}\}, \{o^{(t-h)}, \dots, o^{(t)}\})$$
\end{linenomath*}
, where $h$ is the measurement history taken into consideration. The model output was to approximate the gold label 
\begin{linenomath*}
$$Y \in \{0, 1\}$$
\end{linenomath*}, where 0 represents a non-incident label and 1 represents an incident.


\begin{figure}
    \centering
    \includegraphics[width=\linewidth]{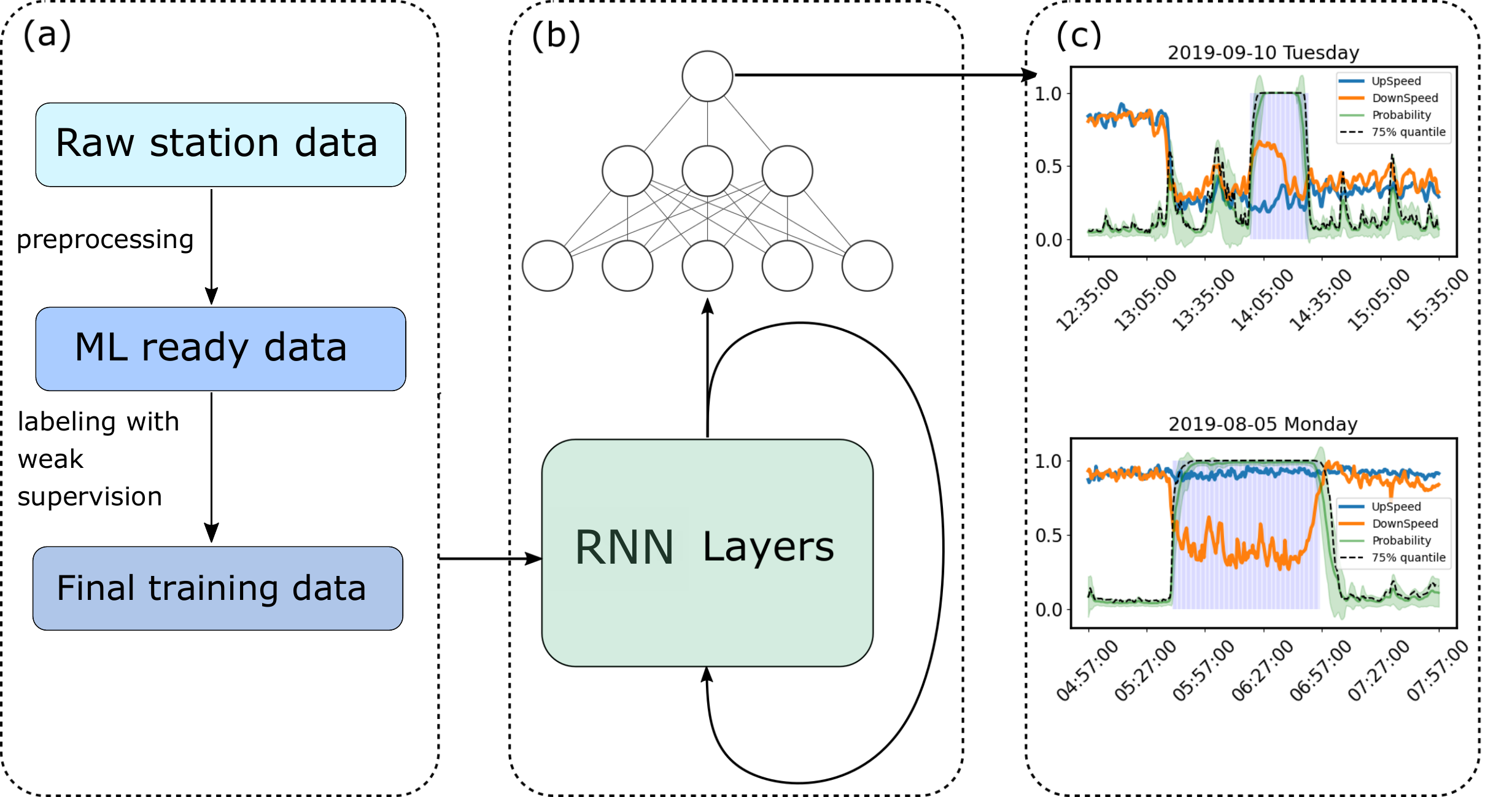}
    \caption{RNN-based incident detection workflow. (a)  Data preprocessing and curation processes, in which the raw station data are first made machine learning ready and then labeled via weak supervision. (b) Data flow in the neural network. The time series data were fed into RNN layers and then into fully connected layers for classification. (c) Process of the model classifying time slices and detection being made.} 
    \label{workflow}
\end{figure}

\subsection{Data curation}
The two datasets described in the preceding section provide the foundation for the approach: the inductive loop measurements that provide the speed profiles at specified locations on the road network, and the associated labeled incident data. Field data were cleaned and smoothed to provide quality data; however, the labeled incident data had significant issues. Challenges included incorrect incident start and end times, as well as incident location overlaps in which a detector is capturing the dynamics of multiple events.

A traffic-affecting incident profile is expected to have a drop on the upstream speed, while maintaining the downstream speed, and an increase in the upstream occupancy. For each timestamp, two outcomes are possible---it belongs to an incident or not. However, the impact of an incident on speed profiles can be diverse. Many times when an incident happens, it has minimal influence on the traffic, for example, an incident that occurs at night  when  fewer vehicles are on the road.  
In this work we focus only on detecting incidents that are sensor detectable, meaning that there was an observed difference between upstream and downstream speeds. The majority of these cases presented a significant drop in the upstream speed and unchanged downstream speed. Road conditions can be complicated, however, and unexpected scenarios can happen. For example, several cases showed upstream and downstream separation where the downstream speed was much lower. Also, in some cases  both upstream and downstream speed profiles had an abrupt drop during an incident. These cases showed a steeper drop and tended to last for a shorter period of time than peak-hour-related traffic profiles. All the sensor-detectable cases described  were selected from the original training set for model training. 

\begin{figure}
    \centering
    \includegraphics[width=0.8\linewidth]{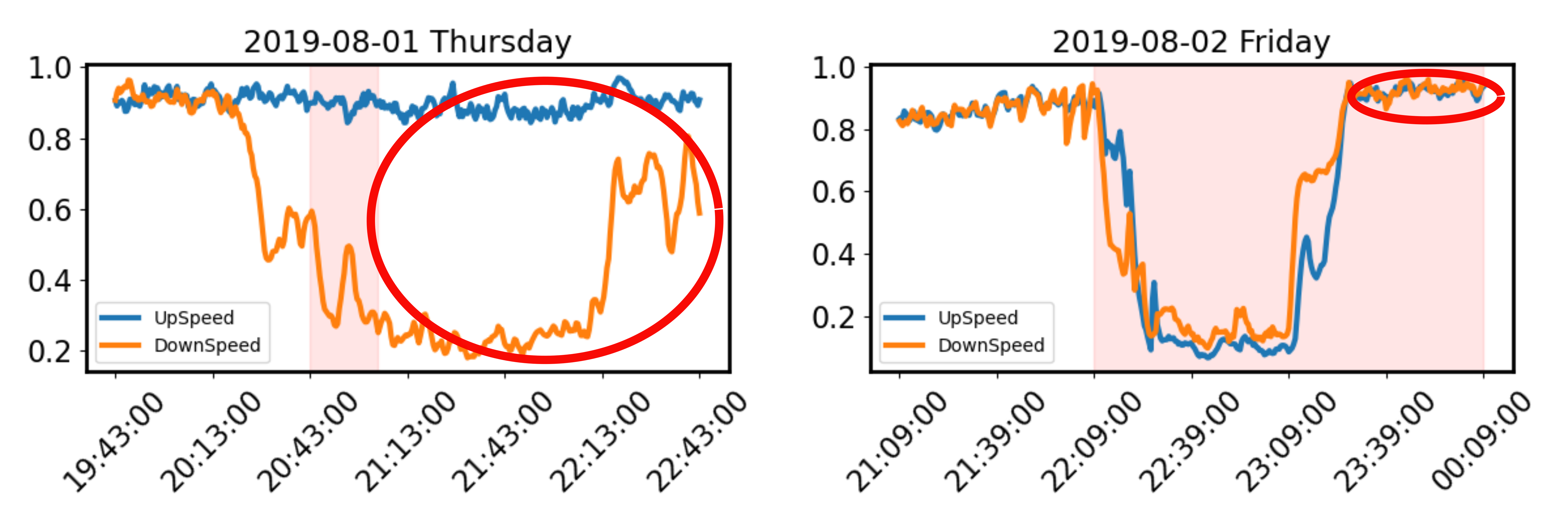}
    \caption{Mislabeled time slices with traffic profiles. The red background indicates the reported accident duration. The speed profiles of the circled areas do not match the reported incidents.}
    \label{miss-label}
\end{figure}

As mentioned, the labeling associated with the sensor-detecable cases was not accurate. A recurring inaccuracy was that the reported incident starting time and duration did not match the traffic profiles. Figure \ref{miss-label} shows an example of poorly labeled sensor data. Using the incident reports directly with the original labels would greatly confuse the models and lead to poor classification performance.

\paragraph{Creating training set with weak supervision}

\begin{figure*}
    \centering
    \includegraphics[width=\linewidth]{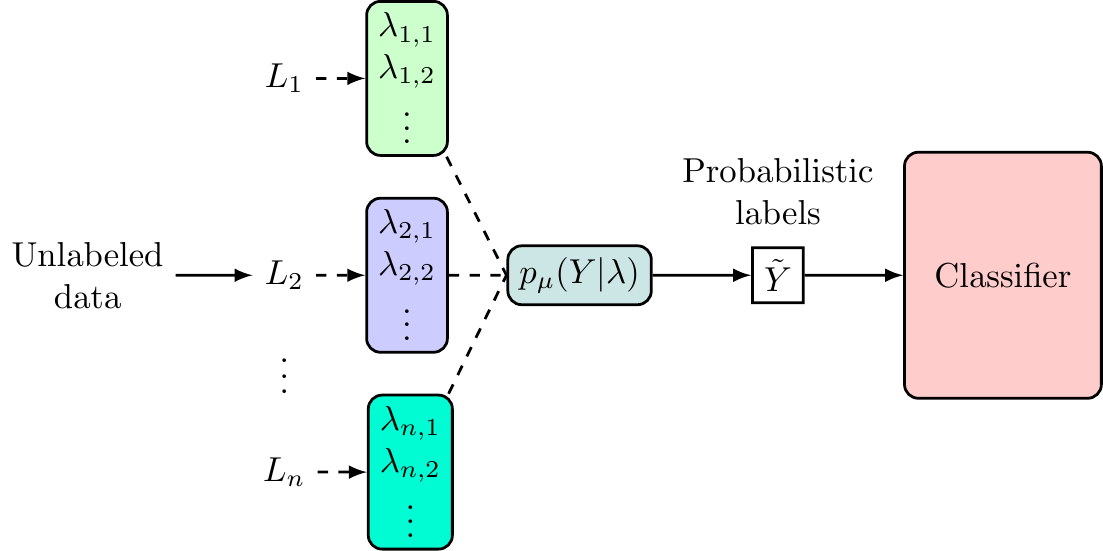}
    \caption{Probabilistic labels creation process using Snorkel. The unlabeled data points get labeled by a set of labeling functions $L_1, \dots, L_n$. The $i$th labeling function produces a label $\lambda_{i, j}$ or abstain for data point $j$. Snorkel then evaluates the labeling function accuracies without accessing the true label and produces probabilistic labels, which can be used directly in the downstream classifier.}
    \label{snorkel}
\end{figure*}

 Weak supervision uses the information from different weak labelers to create high quantity and quality of training data. These labelers can be noisy and imprecise, however, and usually follow a set of rules that are manually determined. In this work we adopt \texttt{Snorkel} \cite{ratner2017snorkel}, a system for quickly generating training data with weak supervision, to label our training set based on a set of labeling functions (LFs). The LFs determine whether a particular data point should be labeled as incident, non-incident, or abstain based on individual criteria. \texttt{Snorkel} produces labeling statistics from all LFs and then leverages the importance of each LF by inspecting the coverage and conflicts of all LFs. It then outputs probabilistic labels for the labeled data points in the training set, encoding the noise brought by the LFs. In essence, \texttt{Snorkel} models the joint probability distribution \begin{linenomath*}
$$p(\lambda, Y),$$
 \end{linenomath*} where $\lambda$ is a vector containing the results from all LFs and $Y$ is the gold label (ground truth), and outputs the conditional probability $p(Y | \lambda)$ as the probabilistic labels for the data points, as shown in Figure \ref{snorkel}. \texttt{Snorkel} is able to quickly model the joint distribution and estimate the accuracy of the LFs without knowing the gold label $Y$, which is done via solving a matrix-completion-style problem \cite{ratner2019training}. More specifically,  a user can define a source graph $G_{source}$ capturing the dependencies between LFs. If no such graph is provided, \texttt{Snorkel} will treat LFs independently conditioned under the true labels and learn the correlation during the process. To model $p(Y|\lambda)$, \texttt{Snorkel} parameterizes the distribution with a vector of source (LFs) correlations and accuracies $\mu = \mathbb{E}[\psi(C)]$,
 where $\psi(C) = \{0, 1\}^{\prod_{i\in \mathcal{C}}(|\mathcal{Y}_i| - 1)}$, $C \in \mathcal{C}$, $\mathcal{C}$ is the set of cliques of $G_{source}$, and $\mathcal{Y}_i$ is the output space from the $i$th source. \texttt{Snorkel} estimates $\mu$ without  access to the true labels, by analyzing the covariance matrix of the cliques in $G_{source}$. The set of the cliques in $G_{souces}$ can be written as $R \cup Q$, where $R$ is the observable cliques and $Q$ is the separator set of cliques. 
 of the junction tree of $G_{source}$. Thus, the random indicator variables $\psi(C) = \psi( R \cup Q)$, and 
 \begin{linenomath*}
$$\mathbf{Cov}[\psi(R \cup Q)] = \begin{bmatrix}
	\Sigma_{R} & \Sigma_{RQ}\\
	\Sigma^T_{RQ} & \Sigma{Q},
\end{bmatrix}$$
which has the inverse
$$K = \mathbf{Cov}[\psi(R \cup Q)]^{-1}=\begin{bmatrix}
	K_{R} & K_{RQ}\\
	K^T_{RQ} & K_{Q}
\end{bmatrix}.
$$
 \end{linenomath*}
 
Here, $\Sigma_R$ is known, and $\Sigma_S$ is known or can be estimated. The source correlations and accuracies $\mu$ can be recovered by obtaining $\Sigma_{RS}$. One can show that if  $z = \sqrt{c}\Sigma_{R}^{-1}\Sigma_{RQ}$, where $c = (\Sigma_Q -\Sigma^T_{RQ}\Sigma_R^{-1}\Sigma_{RQ})^{-1}$, then
\begin{linenomath*}
$$K_R = \Sigma_R^{-1} + zz^T$$
\end{linenomath*}
in which $K_O$ is determined by $G_{source}$ , $\Sigma_R^{-1}$ is observable, and $zz^T$ can directly solve for $\mu$ via an algorithmic approach that treats estimating $z$ as a matrix completion problem. More details of the development and analysis of this method are in the original work \cite{ratner2019training}.

\begin{figure}[t!]
\centering
\includegraphics[width=\linewidth]{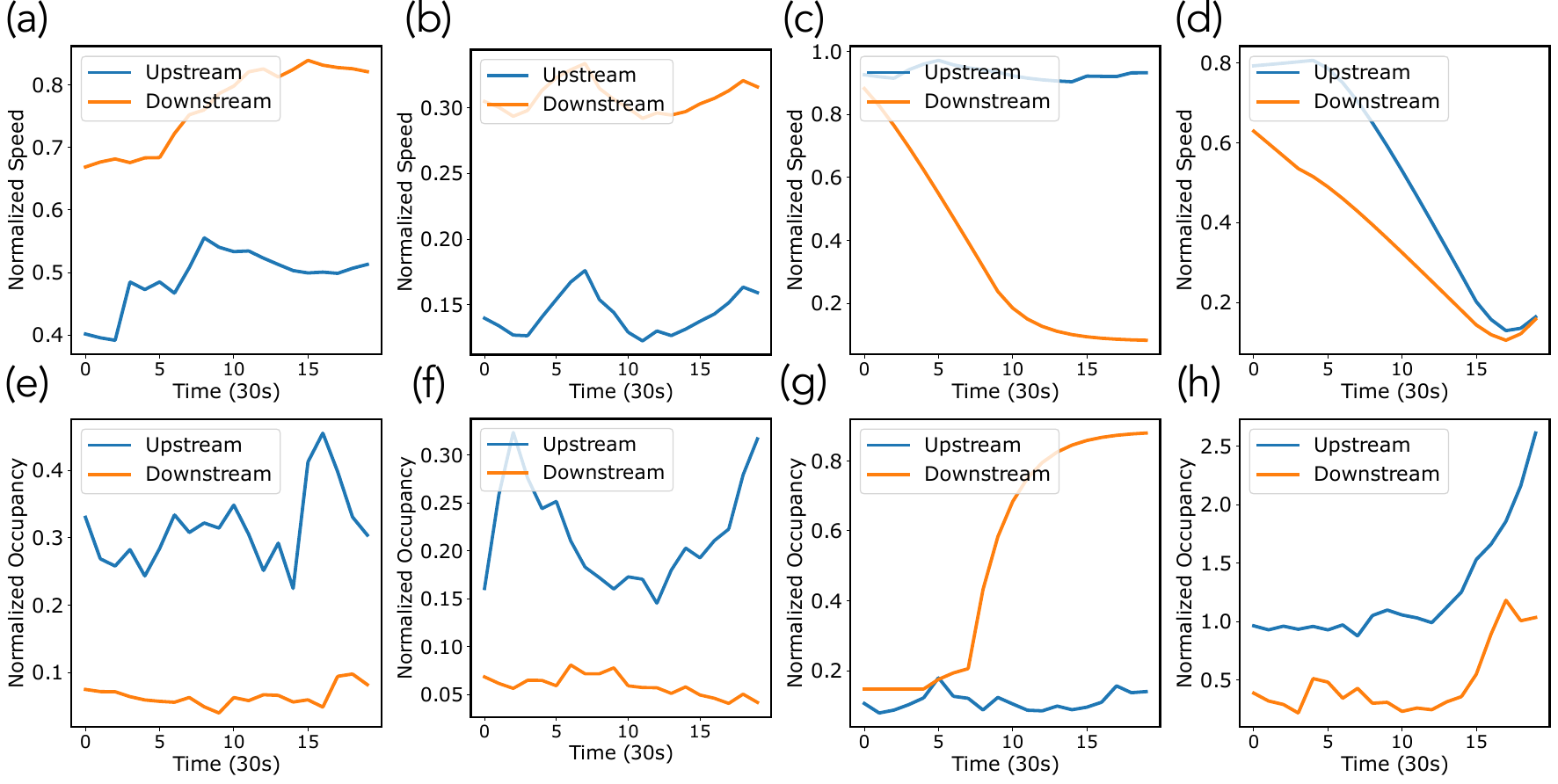}
\caption{Incident examples based on labeling function results for Snorkel. (a)--(d) show the speed-based labeling function results. Specifically, (a) shows upstream and downstream separation; (b) shows more significant separation; (c) presents one-sided speed drop; and (d) shows double-sided speed drop. Similarly, (e)--(f) show the labeling function results based on occupancy. 
(g) shows single-sided occupancy increase, and (h) shows the double-sided occupancy increase. }
\label{LFs}
\end{figure}

For traffic incident detection, given the observations shown in Figure \ref{pattern}, we designed 10 LFs 
related to speed and occupancy as the weak labels to create the training set with probabilistic labels. The LFs are based on speed and occupancy changes (separations, single- or double-sided drop or increase) during an incident. Examples of LF function results with incident labels are shown in Figure \ref{LFs}. In total, 157,430 time slices were labeled by \texttt{Snorkel}, among which 118,054 were non-incident slices and 39,376 were incident slices. 

\begin{figure}
    \centering
    \includegraphics[width=0.7\linewidth]{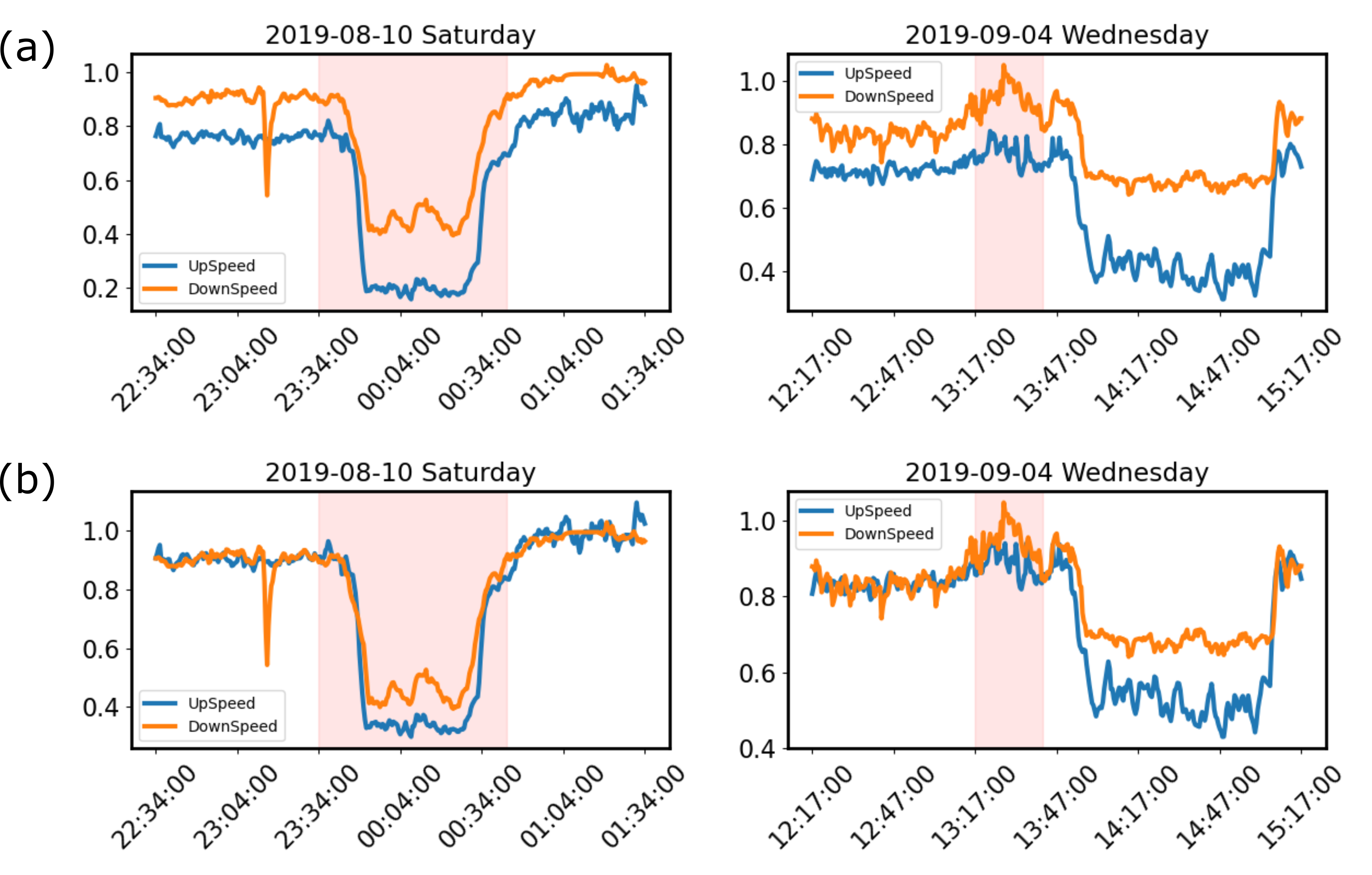}
    \caption{Offset of the upstream speed with one-hour average speed difference during normal traffic. (a) Original speed profiles; (b) profiles with upstream offset.}
    \label{offset}
\end{figure}

\paragraph{Offset upstream speed}
Manually selected non-incident data points had non-separable or close to non-separable upstream and downstream speed profiles. In reality, because of complex road conditions and non-uniform distances between upstream and downstream loop detectors, a constant separation may occur between the speed profiles, shown in Figure \ref{offset}. The models will perform poorly on the non-incident time slices in these cases. To address this issue, we offset the upstream speeds  by the one-hour average speed difference of the downstream speeds, effectively normalizing the speeds.

\subsection{Time slice classification}
A recurrent neural network (RNN) classifier with long short-term  memory  
cells was adopted because of their success and popularity in sequential learning tasks. 
The LSTM network acts as the default model for classifying individual time slices into incident and non-incident classes.  For the AID task, it is important to capture not only the instantaneous behavior of the traffic profiles but also the changing trend of the profiles with time. In this implementation, the LSTM layers take into account the information from the history (10-minute history in the default setting) for each speed profile and return the final state for classification. Classification was achieved by the final dense layers. With the sigmoid activation at the final layer, the  network  outputs the probability values of each time slice being an incident time slice.

The loss function of choice for model training was focal loss (FL) \cite{lin2017focal} due to the class imbalance nature of the problem. It is defined as follows: 
\begin{linenomath*}
$$FL = -(1 - p_t)^{\gamma}log(p_t)$$
\end{linenomath*}, where $p_t$ is the true label probability. FL puts more importance on the harder-to-learn examples that usually are the data points with minor class labels. The parameter $\gamma$ was set to be the default value from its original work \cite{lin2017focal}. We found FL could suffice to serve as both the loss function and the hyper-parameter search (HPS) objective in this work. The workflow of the data-centric RNN-based incident detection is shown in Figure \ref{workflow}.

Ideally, a DR of 1.0 and an FAR of 0.0 indicate a perfect incident detection model. Usually, however, there is  a trade-off between DR and FAR, where a high DR is accompanied by a high FAR. The reason is that a model tends to overdetect in order to keep up with the high detection rate, and vice versa. Thus, the objective of an incident detection model is to extract the hidden traffic incident patterns and increase DR, reducing FAR to the largest extent.

The detection model in our work is time slices, classification based with additional detection criteria described in Section \ref{results}. 
The final DR and FAR are directly related to the classifier performance. The metric used to evaluate the performance of classifiers is the F1 score, which is the harmonic mean of precision and recall. 
Similar to DR and FAR, there is usually a trade-off between precision and recall, where a higher-valued one is accompanied by a lower-valued one. In this work, high values for both precision and recall are desired. Hence, the F1 score $(\frac{2(\text{Precision} \times \text{Recall})}{\text{Precision} + \text{Recall}})$ was utilized to evaluate classifiers, being able to combine both and weigh their contributions.

\subsection{Uncertainty quantification using deep ensemble}

Measuring the predictive uncertainty is critical for using the incident detection model in practice. Uncertainty estimation helps one understand when to trust the model prediction. Bayesian neural networks are used widely to measure the predictive uncertainty. These methods are computationally expensive, however,  and require significant modification of the training procedure of the existing network. Therefore, we used an alternative approach called deep ensemble \cite{lakshminarayanan2016simple} to estimate the predictive uncertainty of our incident detection model. Deep ensemble is capable of estimating high-quality predictive uncertainty. This method is easy to implement and does not require much hyperparameter tuning. 

We used a randomization-based ensemble strategy to train an $M$ LSTM network with various initializations. The parameters of the LSTM model were randomly initialized, and the model was trained on the whole training dataset. We analyzed networks with accuracy greater than a particular threshold and treated their results as a combination of Gaussian distributions to estimate the final prediction from the ensemble predictions. We then calculated the mean and variance of Gaussian distributions using the following formula:


    $$\mu_*(x) = \frac{1}{M} \sum_{i=1}^M \mu_i (\mathbf{x})$$

    $$\sigma_*^2(x) = \frac{1}{M} \sum_{i=1}^M (\mu_i (\mathbf{x}) - \mu_* (\mathbf{x}))^2 .$$

By training the model with 50 different random seeds, selecting well-performing models (classification accuracy $>$ 0.9), and then using the mean and standard deviation of predictions from those models, a statistical measure of uncertainty can be obtained.


\section{Experiment results}
\label{results}
The following summaries the process we followed.
\begin{itemize}
    \item Downloaded PeMS data  for the entire year of 2019 and for January, February, and March 2020. A smaller region on I80 was selected for analysis.
    \item Created the training set using \texttt{Snorkel} with flow profiles (upstream and downstream speed and occupancy) from January and March 2019.
    \item Created the validation set of manually selected profiles from April and May  2019.
    \item Selected sensor-detectable sequences from June to December 2019 as the testing set.
    \item Visually identified true cases of an incident to account for mislabeling in PeMS reported incident data.
\end{itemize}

This process generated the initializing data frames for our \texttt{TensorFlow} implementation. 

The next step was to conduct  a hyperparameter search  for the recurrent neural-network-based models for incident detection. We placed a set of network architecture-related and training process-related hyperparameters into the search space. The network architecture-related parameters were RNN cell type, number of RNN layers, number of units in each RNN layer, whether it is a bidirectional RNN cell, number of dense layers, and number of neurons in each dense layer. The training process parameters were batch size, number of epochs, learning rate, sequence length, and patience for early stopping. The objective function for getting the optimal configuration was set to binary cross-entropy. 
These hyperparameters of the RNN 
network were tuned by using \texttt{DeepHyper} \cite{balaprakash2018deephyper}, a scalable neural network hyperparameter tuning package. 
The evaluation of the classification performance was done on the validation set and detection performance on the testing set.

\subsection{Impact of data curation}
In order  to evaluate the efficacy of the data curation proposed in the preceding section, in addition to the proposed LSTM network, random forest, k-nearest neighbors, and support vector machine  ensemble models were trained and evaluated on the data before and after curation. 
These classifiers could use the  same training data while having fundamentally different learning mechanisms. The models were chosen because the KNN/SVM ensemble and random forest models had been used in past studies \cite{xiao2019svm, lin2020automated} for traffic incident detection.  While the RNN-based model could capture the time dependency within the traffic profiles, the other two models would treat the profiles with 10-minute history independently as individual input features. 

The performance of the trained classifiers in terms of classifying time slices in the selected validation set is shown in Table \ref{cls_perf}. 
The results indicate a dramatic performance improvement for all models after the data curation, especially in terms of  recall on the positive (incident) instances and classification accuracy. Among the models in the experiment, the proposed RNN network and random forest model had similar performance with high accuracy, but the RNN network was able to produce better recall on non-incident samples and precision on incident samples, features that are preferred in real-world applications since they would result in a lower false alarm rate. 

\begin{table}[]
    \footnotesize
    \centering
    \caption{Time slice classification model performance on the validation set before and after data curation}
    \begin{tabular}{l|ccccccc}
        \hline
        \hline
         & \multicolumn{3}{c}{Non-incident class} & \multicolumn{3}{c}{Incident class} &  \\
         &  Precision & Recall & F1 & Precision & Recall & F1 & Accuracy \\
         &\\
         \textbf{Before Weak supervision} & \\
         \hline
         Random Forest & 0.61 & 0.94 & 0.74 & 0.87 & 0.40 & 0.55& 0.67\\
         KNN SVM ensemble & 0.58 & 0.82 & 0.68 & 0.68 & 0.40 & 0.50 & 0.61\\
         LSTM & 0.70 & 0.92 & 0.79 & 0.87 & 0.58 & 0.69 & 0.75\\
         \hline
         \vspace{2em}\\
         
         \textbf{After Weak supervision}& \\
         \hline
         Random Forest & 0.83 & 0.75 & 0.79 & 0.77 & 0.85 & 0.81 & 0.80\\
         KNN SVM ensemble & 0.81 & 0.78 & 0.79 & 0.78 & 0.82 & 0.80 & 0.80\\
         LSTM & 0.92 & 0.83 & 0.87 & 0.83 & 0.92 & 0.88 & 0.87\\
         
    \end{tabular}
    
    \label{cls_perf}
\end{table}

\subsection{Incident detection}\label{detection}
\subsubsection{Metrics}

\begin{equation}
    DR = \frac{\text{Number of correctly detected incidents}}{\text{Number of total incidents}}
    \label{dr}
\end{equation}

\begin{equation}
    FAR = \frac{\text{Number of falsely detected incidents}}{\text{Number of detected incidents}}
    \label{far}
\end{equation}

The performance of a traffic incident detection model can be evaluated via two common metrics: detection rate  and false alarm rate. Detection rate represents the ratio of the number of correctly detected incidents to the number of ground truth incidents. It reflects the detection model's ability to capture potential characteristics of an incident.  The detection rate is  shown in Equation \ref{dr}. However, definitions of detection rate were not consistent in the literature. In \cite{el2018fuzzy} the numerator in the equation appeared to be the number of detected incidents. This is problematic because the number of detected incidents could potentially exceed the total number of incidents when the model makes many false detections. 

On the other hand, the false alarm rate is  the proportion of falsely detected incidents among all the detected incidents. The false alarm rate is defined in Equation \ref{far}. The inconsistency in the literature definitions also apply to FAR, where the number of non-incident cases was in the denominator  \cite{xiao2019svm, zhu2018deep}. This would lead to a low FAR since the number of non-incident cases is usually much larger than that of incidents. 

Therefore, we use the DR and FAR define in the Equations \ref{dr} and \ref{far} for the evaluation of the results in the rest of the our paper.


\subsubsection{Detection Results}
Here we present the incident detection results after applying the trained RNN with the  default LSTM cell and the deep ensemble model to the testing set. The original reported labels for the time slices in the testing set were noisy, with a considerable number of labels not matching the time slices' true behavior. Therefore, we report only the detection metric scores (DR, FAR) on 211 human-detectable incident cases from the testing set. The confirmation of when incidents happened was done visually. Our criterion was to look for incident indicators (speed drop, occupancy increase, and speed and occupancy separation) visually and compare them with the corresponding model detection results. DR and FAR were calculated based on the matches between the model detection and visual investigation. 
The model detection was based on the time slice classification results with two additional detection filters: (1) the time slices classified as incident with free-flow (normalized $>0.8$) speed on both upstream and downstream would be determined as non-incident since the traffic was not affected,  and (2) the number of consecutive time slices classified as incident had to exceed 6 steps (3 minutes) to be considered a detected incident. The LSTM network and deep ensemble model produced probability output for each instance in the testing set, and the final class labels were assigned based on the output threshold of 0.5 ( $>$ 0.5 incident class; $\leq$ 0.5 non-incident class). 


The trained RNN with the default configurations made 231 incident detections, among which 203 were correctly detected and 28 were false alarms. It obtained a DR of 0.96 and an FAR of 0.12. The deep ensemble model was obtained after HPS for the default network and training using 50 different random seeds with the best configuration. Among the 50 models, 25 were selected based on their performance on the validation set (classification accuracy $>$ 0.9) to form the ensemble.  Given the uncertainty measure the deep ensemble model provided, we used a 75\% quantile of the probability outputs to assign labels. 

Figure \ref{detection} shows the detection examples from the deep ensemble model. The deep ensemble model was able to accurately detect the more obvious incidents, having low uncertainty at obvious regions (second plot in Figure \ref{detection}(a)), and it was improved by using the uncertainty measure. The use of the 75\% quantile of probability outputs improved detection on cases similar to the first plot in Figure \ref{detection}(b). However, the model tended to fail to detect incidents when the speed separation between the upstream and downstream was minor and when both speed profiles dropped abruptly but without any separation. 
When the model failed to detect the incidents, it  usually had larger uncertainties. 


\begin{figure}[t!]
    \centering
    \includegraphics[width=0.9\linewidth]{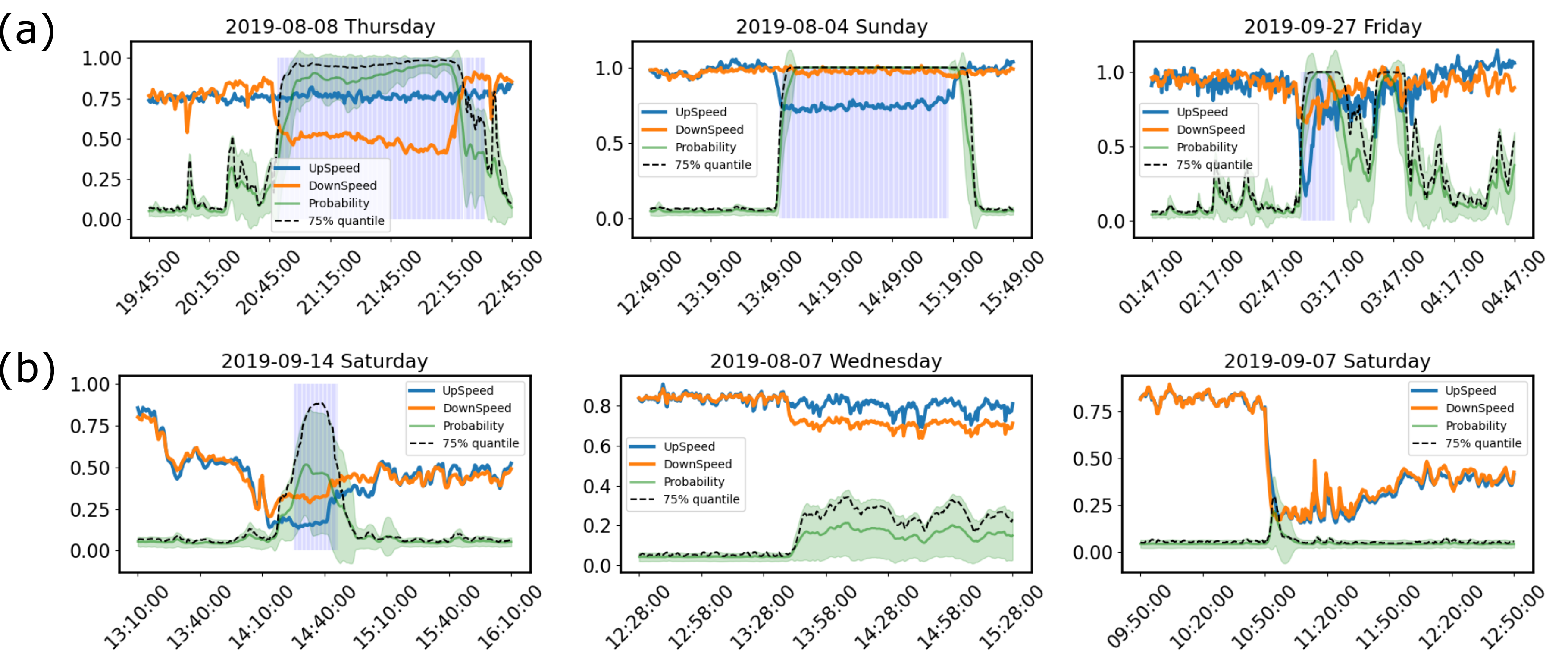}
    \caption{Detection results from the deep ensemble with a 75\% quantile of the probability outputs. The blue highlighted parts indicate detected incidents. (a) Examples with easy-to-detect incidents where the speed separation during incident was evident. (b) Harder examples---the first one demonstrates the advantage of using the 75\% quantile for classification where it successfully detected the incident while the mean values were not exceeding the threshold; the second plot implies that  incidents with minor speed separation might be hard for the model to detect; the third one shows the most easily missed cases where both the upstream and downstream speed had an abrupt drop while there was no speed separation.}
    \label{detection}
\end{figure}

Overall, the deep ensemble model with the 75\% quantile of probability output achieved 0.90 DR and 0.08 FAR with 168 correctly detected incidents and 12 false alarms. Compared with the RNN-based model with default configuration, the deep ensemble model had lower DR and  lower FAR. This result was expected from the model evaluation on the validation set, in which the deep ensemble model had higher precision and lower recall on the positive (incident) data points while keeping the overall classification accuracy the same.

\subsection{Implementing real-time detection}
Our testing process highly resembled real-time detection. The time slices with the speed profile history were fed to the model, the model classified these slices, and then detection was made based on additional criteria. We envision a similar process for the model to work in real time. These steps are described in more detail below.
\begin{itemize}
    \item Obtain one-hour average upstream and downstream speed difference during normal traffic. The one-hour window should be as close to the detection period as possible. This can be done via a moving average with a one-hour window excluding the contribution brought by incident-related profiles.
    \item Stream time slices with speed profile history; offset the upstream speed with the one-hour average difference; obtain relative speed difference.
    \item Feed processed data to the model; get model predictions with the incident or non-incident labels.
    \item Raise ``require attention" flag when the coming time slices are classified as incident slices.
    \item Raise ``incident detected" flag when there are more than 6 consecutive ``require attention" flags.
\end{itemize}

\begin{figure}[t!]
    \centering
    \includegraphics[width=\linewidth]{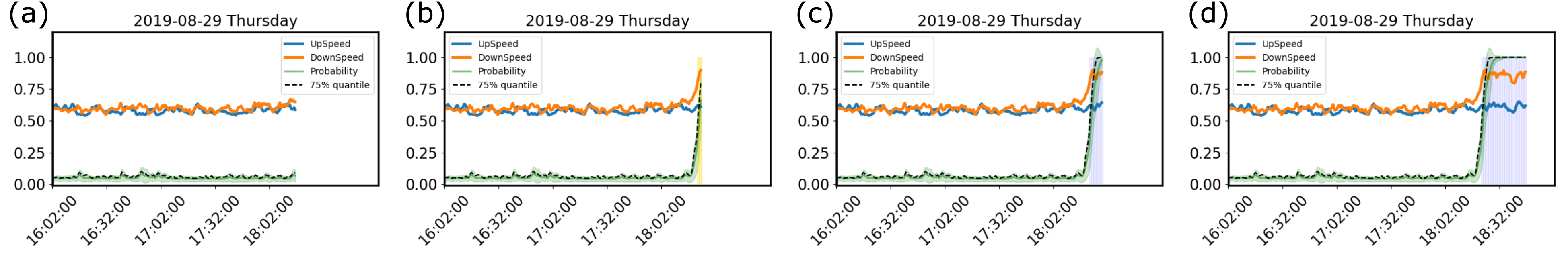}
    \caption{Real-time incident detection process using the deep ensemble model. (a)  Normal traffic -- model output probability is low with low uncertainty; (b) ``Require attention" status -- model output probability exceeded the threshold but the consecutive number of classified incident slices did not meet incident status; (c) Status changed to ``incident detected" after the number of consecutive classified incident slices went over 6; (d) Detection results for additional time.}
    \label{realtime}
\end{figure}

Figure \ref{realtime} shows the process of real-time detection. This process is robust to the possible noise present in the loop detector measurements. In particular, the ``require attention" status can act as a buffer for determining whether the classified incident slices are resulting from faulty measurements. Consequently, the detection is bound to be at least 6 steps (3 minutes) later than the actual starting time of the incidents.

\section{Conclusion and future work}
\label{con}

We presented a data-centric supervised learning workflow for  sensor level traffic incident detection. The training and validation data quality were ensured via preprocessing (exponential smoothing; oscillation check) and curation (input-label pair check; data selection; upstream speed offset). Demonstration with  three different classification models, including the proposed RNN-based network, showed that models trained on the carefully curated data had enormous classification performance improvement. In addition, the proposed RNN-based network (LSTM cell by default) was tuned through a hyperparameter search. The tuned model configurations were used for creating the deep ensemble model, which provides quantified predictive uncertainties. The detection results on 211 sensor detectable cases in the testzving set show that for both the default RNN model and the deep ensemble model we can achieve a high detection rate (0.96; 0.90) and low false alarm rate (0.12; 0.08). A real-time incident detection process using our RNN-based model was suggested, in which three detection statuses would be returned from the process.

We point out a few limitations of the current work and suggest future work to address these limitations. (1) The proposed incident detection method greatly depends on the traffic dynamics where the stop-and-go conditions in rush hours resemble incident behavior, resulting in false alarms. For future work, we plan to incorporate traffic wave models to reduce false detection further.
(2) The uncertainty measure provided by the deep ensemble model was not fully exploited. Our future work will involve evaluating different statistics of the output probability values to classify instances and creating protocols for making decisions at the regions with high predictive uncertainties. (3) Only speed and occupancy information from the loop detector measurements was used. Besides other loop detector measurements,  many other data sources, such as surveillance video feed and vehicle GPS information,  can be helpful for detecting incidents. We aim to combine information from different sources to develop incident detection models for data fusion to further improve detection accuracy.


\section*{Acknowledgments}
This material is based in part upon work supported by the U.S. Department of Energy, Office of Science, under contract DE-AC02-06CH11357. 
This research used resources of the Argonne 
Leadership Computing Facility, which is a DOE Office of Science User Facility under contract DE-AC02-06CH11357. 
This report and the work were sponsored by the U.S. Department of Energy (DOE) Vehicle Technologies Office (VTO) under the Big Data Solutions for Mobility Program, an initiative of the Energy Efficient Mobility Systems (EEMS) Program. David Anderson and Prasad Gupte, the DOE Office of Energy Efficiency and Renewable Energy (EERE) managers played important roles in establishing the project concept, advancing implementation, and providing ongoing guidance.

\bibliography{references}

\bibliographystyle{unsrt}





\clearpage
\pagestyle{empty} 
\begin{figure*}
\begin{center}
    \framebox{\parbox{6in}{
    The submitted manuscript has been created by UChicago Argonne, LLC, Operator of Argonne National Laboratory (``Argonne''). Argonne, a U.S. Department of Energy Office of Science laboratory, is operated under Contract No. DE-AC02-06CH11357. The U.S. Government retains for itself, and others acting on its behalf, a paid-up nonexclusive, irrevocable worldwide license in said article to reproduce, prepare derivative works, distribute copies to the public, and perform publicly and display publicly, by or on behalf of the Government. The Department of Energy will provide public access to these results of federally sponsored research in accordance with the DOE Public Access Plan. \url{http://energy.gov/downloads/doe-public-access-plan}}}
    \normalsize
\end{center}
\end{figure*}

\end{document}